\documentclass[a4paper]{llncs}

\usepackage{amstext}
\usepackage{amsmath}
\usepackage{amssymb}
\usepackage{url}
\usepackage{enumerate}
\usepackage{graphicx}
\usepackage{makecell}
\usepackage{listings}
\usepackage{wrapfig}
\usepackage{paralist}
\usepackage{xspace}
\usepackage{color}
\usepackage{times}
\usepackage{proof}
\usepackage{algorithm}
\usepackage[noend]{algpseudocode}
\usepackage{color}
\usepackage{capt-of}
\usepackage{longtable}
\usepackage{multirow}
\usepackage[table]{xcolor}

\usepackage{subfigure}
\usepackage[colorinlistoftodos,bordercolor=white]{todonotes}

\title{A hybrid controller for safe and efficient collision avoidance control
}

\author{
    Qiang Wang\inst{1}
   \and Xinlei Zheng\inst{2}
   \and Jiyong Zhang\inst{2}
   \and Joseph Sifakis\inst{1,3}
 }

\institute{
 Research Institute of Trustworthy Autonomous Systems, SUSTech , Shenzhen, China 
\and School of Automation, Hangzhou Dianzi University, Hangzhou, China 
\and Verimag, Universit\'e Grenoble Alpes, France
}

\begin{document}

\maketitle

\begin{abstract}
We design and experimentally evaluate a hybrid safe-by-construction collision avoidance controller for autonomous vehicles. 
 The controller combines into a single architecture the respective advantages of an adaptive controller and a discrete safe controller. 
 The adaptive controller relies on model predictive control to achieve optimal efficiency in nominal conditions. 
 The safe controller avoids collision by applying two different policies, for nominal and out-of-nominal conditions, respectively. 
 We present design principles for both the adaptive and the safe controller and 
 show how each one can contribute in the hybrid architecture to improve performance, 
 road occupancy and passenger comfort while preserving safety. 
 The experimental results confirm the feasibility of the approach and the practical relevance of hybrid controllers for safe and efficient driving.

\end{abstract}

\section{Introduction}

It is widely believed that the deployment of autonomous vehicles can improve not only the traffic efficiency, but also its safety.
 Collision avoidance, as a fundamental safety requirement for autonomous vehicle control,
 plays a crucial role in guaranteeing traffic safety and reducing the number of vehicle crashes,
 given the fact that  more than 50 percent of the total amount of vehicle crashes  are rear-end collisions
 \footnote{http://www.ntsb.gov/safety/safetystudies/Documents/SIR1501.pdf}.

A variety of approaches and frameworks has been investigated for collision avoidance control.
 The underlying assumptions vary largely with the level of modeling of the vehicle dynamics and the nature of the controller stimuli.
 Control-based techniques typically focus on collision avoidance 
 for adaptive cruise control \cite{park2009obstacle,mpc2014} 
 taking into account the impact of perception uncertainty and accuracy of vehicle models \cite{li2013,robust2014}.
 They allow achieving optimality for specific tasks or scenarios without providing strict safety guarantees.
 Model Predictive Control (MPC) \cite{mpc2014,survey2000}, 
 as a prominent optimal control approach has been widely used for vehicle control  
because it allows handling multiple constraints in a receding horizon.
 Nonetheless, MPC relies on the use of optimization algorithms and by its nature cannot guarantee safety, namely collision avoidance.
 Furthermore, depending on the optimization algorithms and the dynamic model of the vehicles, 
 it may also result in high computational complexity because of heavy iterative calculations.

A different line of works focus on safety using formal methods.  
 These apply a variety of techniques including reachability analysis \cite{loos2011adaptive,atva18motionplanner},
 Responsibility-Sensitive Safety (RSS) model \cite{shalev2017formal}, 
 logic-based controller synthesis \cite{nilsson2015correct,sadraddini2017provably},
 as well as the design of safety supervision mechanisms for specific scenarios \cite{korssen2017systematic,krook2019design}.
 The basic principle of safe collision avoidance control,  as formalized and implemented in our previous work \cite{wang2020safe},
 is to keep a safe distance with the preceding vehicles such that in any case 
 the ego vehicle has enough space to brake and avoid collisions.
 Although these results can guarantee correctness by construction,
 they lead to solutions that privilege strict safety at the expense of efficiency.
 Designing a collision avoidance controller for autonomous vehicles 
 that meets both efficiency and safety requirements remains a non-trivial problem.
 The two types of requirements are antagonistic as efficiency implies conflicting properties 
 such as performance, i.e., maximization of the average speed, 
 road occupancy, i.e., keeping the inter-vehicle distance as small as possible 
 and comfort, i.e., no sudden speed changes.

In search of solutions seeking compromises between efficiency and safety, 
 a few works adopt a hybrid approach combining continuous and discrete control dynamics.
 The continuous controller is supervised by an automaton that takes over to handle critical situations.
 Hybrid approaches rely on a principle of "division of roles" often applied in systems engineering 
 that distinguishes between nominal operating conditions and out-of-nominal ones \cite{2018Robust}. 
 The continuous controller has parameters tuned to achieve given goals for nominal operation 
 while the discrete controller deals with out-of-nominal situations.
 For example, \cite{provebecorrectacc,2017Adaptive} propose a switch-control approach,
 where a MPC Controller is safeguarded by an emergency maneuver activated to avoid collision 
 when the ego vehicle is in a critical situation.
 Nonetheless, the emergency maneuver only takes care of safety 
 without investigating possible trade-offs between safety and efficiency of the switch-control policy.

\begin{figure}[h]
\centering
\includegraphics[scale = 0.45]{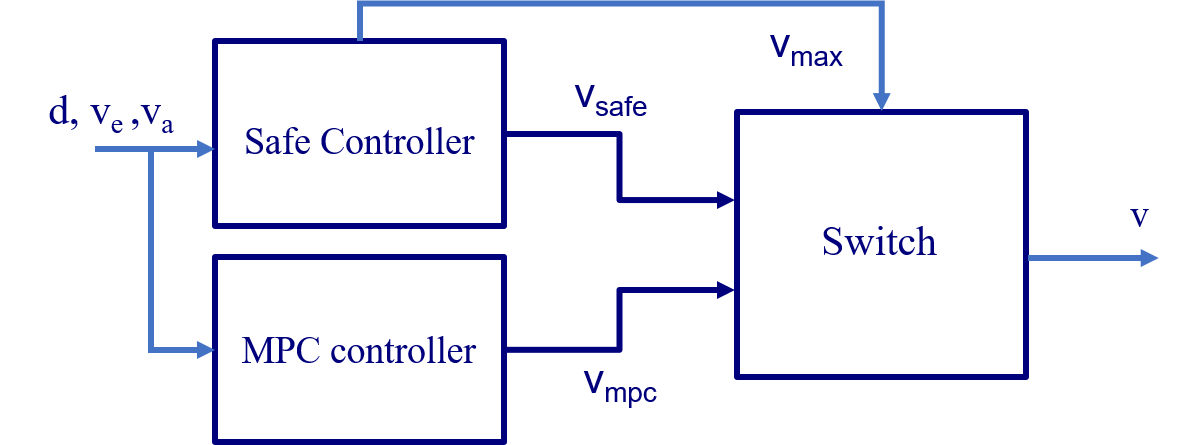}
\caption{The architecture of the Hybrid Controller}
\label{combination-arch}
\end{figure}

We design and experimentally evaluate a safe-by-construction Hybrid Controller 
 that proves to be efficient for the three mentioned criteria. 
 The architecture of the Hybrid Controller is shown in Fig.\ref{combination-arch}.
 It results from the integration into a single architecture of a nominal 
 MPC Controller and the discrete Safe Controller presented in \cite{wang2020safe}. 
 The two controllers running in parallel receive the speed $v_e$ and $v_a$ of the ego vehicle
 and the vehicle ahead respectively, as well as their distance $d$,
 and compute the target speeds $v_{mpc}$ and $v_{safe}$ respectively.
The control policies for computing $v_{mpc}$ and $v_{safe}$ adopt nominal conditions. 
 In particular, $v_{safe}$ is a safe speed under the assumption 
 that the speed of the vehicle ahead is a continuous function and the deceleration does not exceed some limit 
 corresponding to normal driving conditions. 
 In addition to $v_{safe}$, the Safe Controller provides a speed $v_{max}$ 
 that is the maximal safe speed for out-of-nominal conditions when the vehicle ahead suddenly stops, e.g. in case of accident. 
 This speed is computed as a function of the relative distance between the ego vehicle and the car ahead 
 with the maximum deceleration rate of the ego vehicle.
 The Hybrid Controller uses a Switch selecting between the three speeds $v_{mpc}$, $v_{safe}$  and $v_{max}$
 to optimize efficiency criteria while preventing the speed of the ego vehicle to exceed $v_{max}$.
 We show that the combined use of $v_{mpc}$, $v_{safe}$ and $v_{max}$ ensures both efficiency and safety in nominal conditions 
 and moreover safety is preserved in out-of-nominal situations.

Our solution is inspired by the Simplex architecture principle \cite{simplex2001}, 
 for runtime assurance of safety-critical systems. 
 The architecture uses a Decision Module  that switches control from a high-performance 
 but unverified (hence potentially unsafe) Advanced Controller 
 to a verified-safe Baseline Controller if some safety violation is imminent. 
 Nonetheless, in our solution the Safe Controller contributes not only to out-of-nominal situations 
 but also to some nominal situations where it proves to be more efficient than the MPC Controller. 
 Hence, not only the Hybrid Controller is safe but also efficiency gains from the synergistic collaboration are substantial.  
 Of course, the alternation of roles between the MPC and the Safe Controller should be implemented 
 so as to avoid sudden changes of the kinematic state of the vehicle. 
 In particular, care should be taken to avoid jerk (i.e., abrupt changes of acceleration) that might cause passenger discomfort.

Additionally, the paper proposes a pragmatic methodology for the comparative evaluation of the three controllers 
 for two types of scenarios: 
 1) nominal scenarios where the speed of the vehicle ahead is a known continuous function; and 
 2) out-of-nominal scenarios where the vehicle ahead abruptly brakes.

For nominal scenarios, the three controllers are evaluated against three efficiency criteria. 
\begin{itemize}
 \item The first criterion is performance that measures how much close the speed of the ego vehicle can get to the speed of the  vehicle ahead. 
 For a period of time it can be defined as the ratio of the average speed of the ego vehicle with respect to the average speed of the vehicle ahead. 
 This ratio is less than one if the distance between the two vehicles is initially zero.
 \item The second criterion is road occupancy that measures how much close can get the ego vehicle to the vehicle ahead in collision-free scenarios. 
 \item The third criterion measures passenger comfort that decreases as the standard deviation of the acceleration increases.
\end{itemize}

The paper is organized as follows. 
 Section 2 presents the design principles for the MPC Controller and the Safe Controller,
 as well as a comparative study for the two types of scenarios 
 and the three efficiency criteria using the Carla simulator. 
 Section 3 presents the design and the implementation of the Hybrid Controller 
 and its experimental evaluation. 
 Section 4 emphasizes the feasibility and the practical relevance of hybrid controllers 
 for safe and efficient driving and then outlines directions for future work.

\section{A comparative study of the two control approaches}
\label{sec:comparison}

\subsection{The MPC Controller}
 
The MPC paradigm combines three key components.
 The first is a dynamic model of the ego vehicle, 
 allowing the MPC Controller to predict the vehicle states in a given horizon for changing inputs.
 The vehicle state is denoted by the vector $x = [p, v, a]^{T}$, 
 where $p$ is the vehicle position, $v$ is the speed and $a$ is the acceleration.
 Similarly, the state of vehicle ahead is $x_a = [p_a, v_a, a_a]^{T}$.
 The relative distance between the two vehicles is then $d = p_a - p$.
 We further require that the acceleration is buffered as follows:
\begin{equation}
\dot{a} = \frac{u - a}{\tau}
\end{equation}
where the control stimulus $u$ is the desired acceleration, 
 and $\tau$ is the time constant of the actuator lag that captures the inertial characteristics of the vehicle actuator.
 The vehicle dynamics model is described by the following equation.
\begin{equation}
\dot{x} = A_\tau \cdot x + B_\tau \cdot u
\end{equation}
where
\begin{equation}
A_\tau = \begin{bmatrix} 0 & 1 & 0 \\ 0 & 0 & 1 \\ 0 & 0 & -\frac{1}{\tau} \end{bmatrix}, 
B_\tau = \begin{bmatrix} 0 \\ 0 \\ \frac{1}{\tau} \end{bmatrix}
\end{equation}

In order to enhance the stability of the system under the constant time sampling control method,
 we discretize the vehicle dynamics model \cite{dorf2011modern}.
 If $\Delta t$ is the discretization pace,
 the discrete longitudinal dynamics model of the ego vehicle at time instant $t_k$ is given as follows:
\begin{equation}
x(t_{k+1}) = A_{d} \cdot x(t_k) + B_{d} \cdot u (t_k)
\end{equation}
where
\begin{equation}
\begin{aligned}
A_{d} &= e^{A_{\tau} \Delta t} = \begin{bmatrix} 
	1 & \Delta t &  \tau^2 \left(e^{-\frac{\Delta t}{\tau}} - 1 \right) + \Delta t \tau \\
	0 & 1 &             \tau \left(1 - e^{-\frac{\Delta t}{\tau}}\right) \\
	0 & 0 &                       e^{-\frac{\Delta t}{\tau}} \\
\end{bmatrix} \\
B_{d} &= \int_0^{\Delta t} e^{A_{\tau} t} dt \cdot B_{\tau}  = \begin{bmatrix} 
\tau^2 \left(1 - e^{-\frac{\Delta t}{\tau}}\right) + \frac{\Delta t^2}{2} - \Delta t \tau \\
\tau \left(e^{-\frac{\Delta t}{\tau}} - 1 \right) + \Delta t\\
1 - e^{-\frac{\Delta t}{\tau}}
\end{bmatrix}
\end{aligned}
\end{equation}

In addition to the state of the ego vehicle, 
 the MPC Controller also estimates the position of the vehicle ahead in order to compute predictions.
 We assume that in each MPC prediction horizon, the vehicle ahead decelerates at a constant rate $a_{a}$.
 Thus, the position of the vehicle ahead before stopping can be estimated as follows.
\begin{equation}
 p_{a}(t_k) = p_{a}(t_0) + v_{a}(t_0) \cdot (t_k - t_0) + 1/2 \cdot a_{a} \cdot (t_k - t_0)^2 
\end{equation}

The second key component is a cost function, 
 which describes the expected behavior of the ego vehicle, in order to minimize the relative distance.
 Optimization consists in finding the best possible inputs that minimize the cost function.
 The cost function for time horizon $h$ is modeled as a standard quadratic function 
 and the optimization problem is formulated as follows:
\begin{equation}
\label{optimization-equation}
argmin(u^{*}(\cdot), \sum_{k=1}^{h} (x_{opt}^{T} Q x_{opt} + r u^2) ) 
\end{equation}
where $Q$ is the weighting matrix for the state vector and $r$ is the weight for the control stimulus. 
 The new state variable $x_{opt}$ is used to get the relative distance as close as possible to the constant $d_{c}$
 and the speed of the ego vehicle to $v_a$. 
 It is defined by
 \begin{equation}
 \label{x-opt}
 x_{opt} = x_{a} - x - [d_{c}, 0, 0 ]^T
 \end{equation}

The weighting matrix is a diagonal matrix $Q=diag[q_p , q_v , q_a]^T$ , 
 where $q_p$, $q_v$ and $q_a$ are weighting parameters for vehicle position, speed and acceleration respectively. 
 By adjusting the values of these weighting parameters, we can configure preference of  the MPC control tendency.
 For instance, by enlarging the value of $q_p$ we force MPC to drive the ego vehicle closer to the vehicle ahead,
 so that to reduce the relative distance.

Additionally while performing the optimization,
 the MPC Controller enforces the following constraints on the minimum or maximum values of speed and acceleration of the vehicle:
\begin{equation}
\begin{cases}
 v_{min} \leq v \leq v_{max} \\
 u_{min} \leq u \leq u_{max}
\end{cases}
\end{equation}
where $u_{min}$, $u_{max}$ (and $v_{min}$, $v_{max}$) are user-specified parameters for control stimulus and speed, respectively.

The third component of the MPC Controller is the optimization algorithm 
 for solving this linear quadratic programming problem. 
 For this purpose, we use the open source Python library, cvxopt \cite{andersen2011interior}.

\begin{figure} [H]
  \centering
  \includegraphics[width=0.6\hsize]{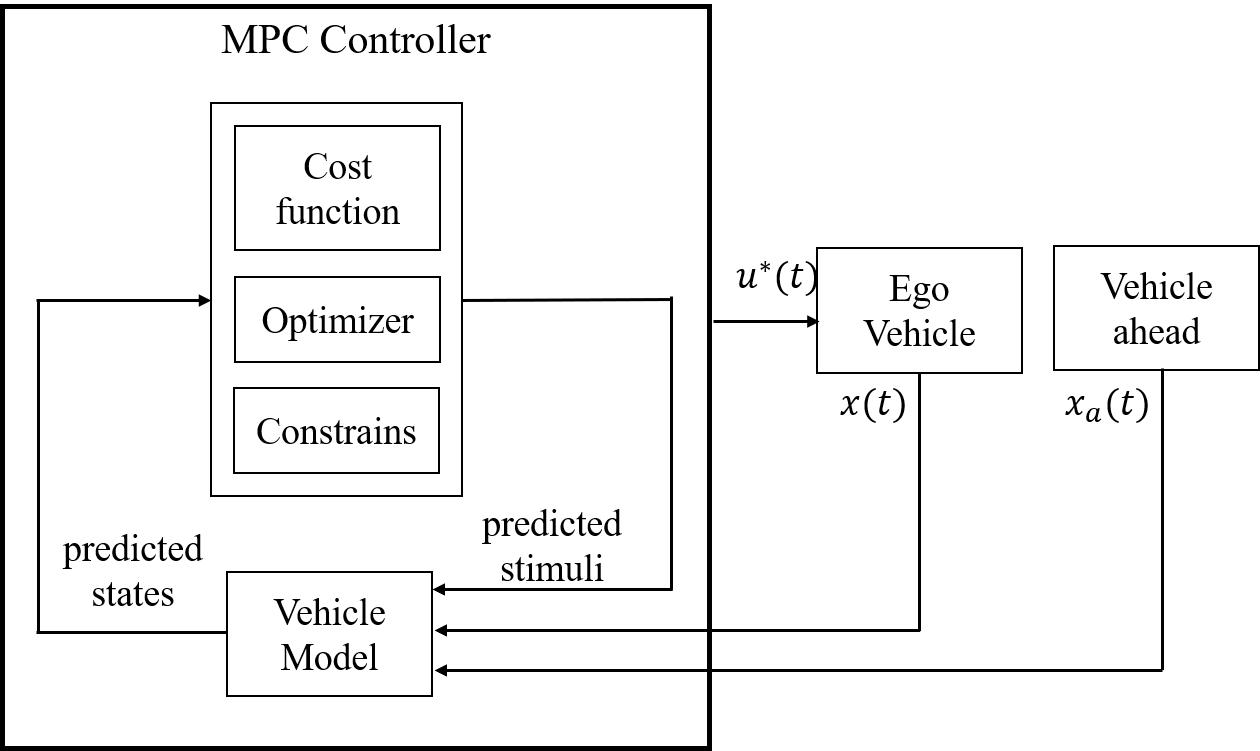}
  \caption{Architecture of the MPC controller}
  \label{mpc}
\end{figure}

Fig.\ref{mpc} shows the architecture of the MPC Controller.
 The controller executes an iterative process optimizing the predictions of vehicle states while manipulating inputs for a given horizon.
 The predictions are based on the specified kinematic model of the vehicle. 
 For each control cycle at time $t_k$, 
 the controller takes as input the current states of the ego vehicle and of the vehicle ahead,
 and computes the future states of the ego vehicle to predict the optimal control stimuli $u^{*}$ 
 minimizing the cost function in the interval [$t_k$, $t_{k + h}$], where $h$ is the prediction horizon.
 The MPC Controller chooses the first element in the sequence as the control stimulus for the ego vehicle,
 and repeats the cycle at time $t_{k + 1}$.
 A key advantage of MPC policy is flexibility in achieving complex objectives and implementing multiple constraints
 when performing optimizations.

\subsection{The Safe Controller}
\label{safe-avoidance}

In a recent work \cite{wang2020safe}, 
 we propose a correct-by-design safe and efficient controller for autonomous vehicles. 
 The controller minimizes the distance between the ego vehicle and the vehicle ahead while preserving safety
 for both nominal and out-of-nominal conditions.

For nominal conditions, the Safe Controller is based on the relative speed $v_{e} - v_{a}$ between the ego vehicle and the vehicle ahead. 
 It computes the target speed $v_{safe}$ for moderate nominal acceleration and deceleration rates to enhance passenger comfort. 
 For out-of-nominal conditions, the Safe Controller computes a target speed $v_{max}$ 
 taking into account only the speed of the ego vehicle and the needed braking distance.
 The braking distance is computed for a maximal deceleration rate that is much larger than the nominal deceleration to cope with dangerous situations, 
 e.g., sudden stops of the vehicle ahead caused by accidents. 
 The controller always keeps the speed $v_{safe} \leq v_{max}$ to make sure that in all circumstances safety is preserved.

We briefly review the general design principle that we specialize for nominal and out-of-nominal conditions. 
 The safe control policy relies on the following three functions.
\begin{itemize}
  \item The function $d(t)$ gives the relative distance at time $t$ between the ego vehicle and the vehicle ahead, 
   which is either stopped or moving in the same direction.
  \item  The braking function $B(v, v')$ gives the distance travelled by the ego vehicle,
  when braking from initial speed $v$ to target speed $v'$.
  When the target speed is $v' = 0$ (i.e, the ego vehicle brakes to a stop),  
  this function is abbreviated as $B(v)$ for simplicity.
  \item  The accelerating function $A(v, v')$ gives the distance travelled by the ego vehicle,
  when accelerating from initial speed $v$ to target speed $v'$.
\end{itemize}

We make no specific assumptions about the implementation of the accelerating and braking functions,  
 e.g., whether acceleration is constant or variable.
 We simply require that the following properties hold:

\begin{itemize}
  \item  $B(v, v') = 0$ and $A(v, v') = 0$ if and only if $v = v'$.
  \item  Additivity property: 
  $B(v, v_1) + B(v_1, v_2) = B(v, v_2)$ and $A(v, v_1) + A(v_1, v_2) = A(v, v_2)$.
  \item  Strict monotonicity:
 $B(v, v_1) < B(v, v_2) $ and $A(v, v_1) < A(v, v_2) $ if $v_1 < v_2$.
\end{itemize}

The basic idea for avoiding collision is to moderate the speed of the vehicle and
 anticipate the changes of the relative distance so as to have enough space and time to adjust and brake.
 For any time $t$, the vehicle only needs to keep track of the distance $d(t)$ and
 check in real-time whether $d(t)$ is greater than the minimal safe braking distance $B(v_t)$ for the current speed $v_t$.
 It starts braking as soon as $d(t)$ reaches the minimal safe braking distance.
 In this way, it is guaranteed that if the obstacles ahead do not move in the opposite direction, no collision would happen.

We consider that the vehicle speed can change between a finite set of increasing levels $v_0, v_1, ..., v_n$,
 where $n$ is a constant, $v_0 = 0$ and $v_n$ is the limit speed of the vehicle.
 The triggering of acceleration and braking from one level to another is controlled according to the free distance ahead 
 and based on the bounds computed as follows,  for each speed level $v_i, i\in [1, n]$,

 \begin{itemize}
   \item $B_{i} = B(v_i)$ is the minimal safe braking distance needed for the vehicle to fully stop from speed $v_i$;
   \item $D_{i} = A(v_{i-1}, v_{i}) + B(v_i)$ is the minimal safe distance needed for the vehicle
 to accelerate from speed $v_{i-1}$ to $v_i$ and then brake from $v_i$ to stop.
 \end{itemize}

We show in \cite{wang2020safe} that the following function specifies the highest safe speed level $v$ 
 as a function of the current speed $v_t$ of the vehicle and the distance $d$, 
 provided that their initial values $v_0$ and $d_0$ satisfy the condition $B(v_0) \leq d_0$.
\begin{equation*}
  v = Control(d, v_t)
\end{equation*}
\begin{equation*}
 v =
    \begin{cases}
    v_{i+1} & \text{when ~~ $v_t = v_i \wedge d = D_{i+1}$} \\
    v_{i-1} &  \text{when ~~ $v_t = v_i \wedge d = B_{i}$} \\
    v_{i} & \text{when ~~ $v_t = v_i \wedge D_{i+1} > d > B_{i}$}
    \end{cases}
\end{equation*}

\begin{figure}[h]
\centering
\includegraphics[width=0.7\textwidth]{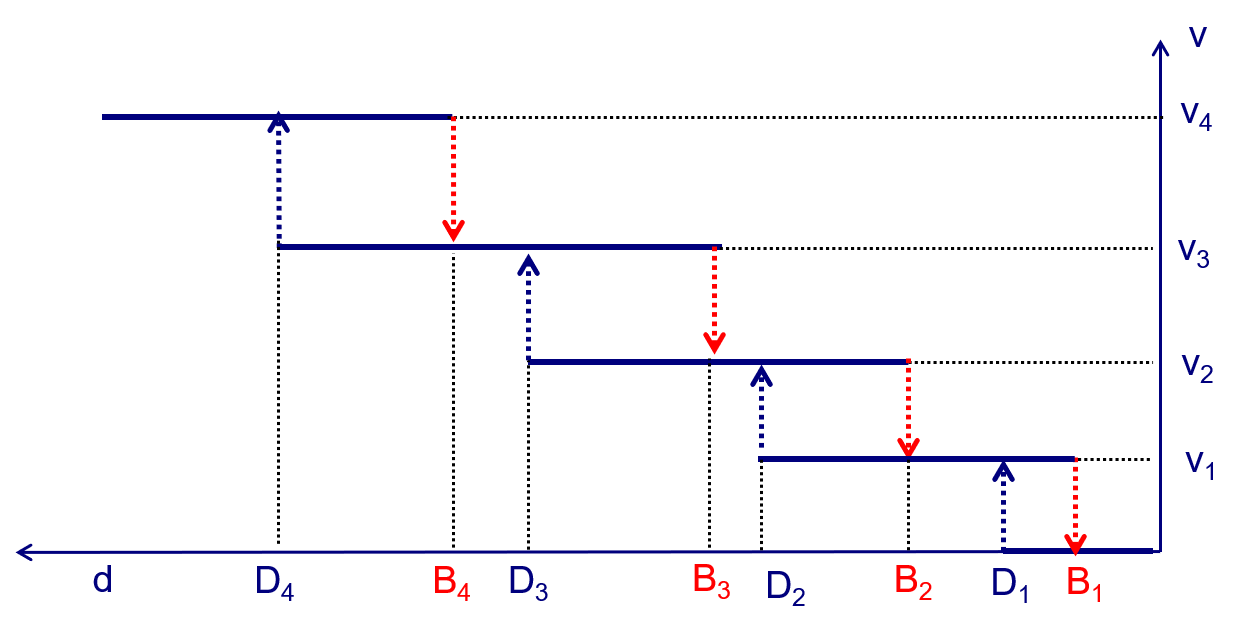}
\caption{Illustration of the collision avoidance principle for $n=4$}
\label{ledder}
\end{figure}

Fig.\ref{ledder} illustrates the principle for $n=4$ speed levels.
 As the value of $d$ increases, the speed of the vehicle climbs up levels.
 Safety is preserved by construction.
 The vehicle can accelerate to a higher level, 
 if it can safely and efficiently use the available distance by combining acceleration and decelaration; 
 in particular braking to a lower level if the distance reaches the bound for safe braking.

\begin{figure}[h]
  \centering
  \includegraphics[width=0.7\textwidth]{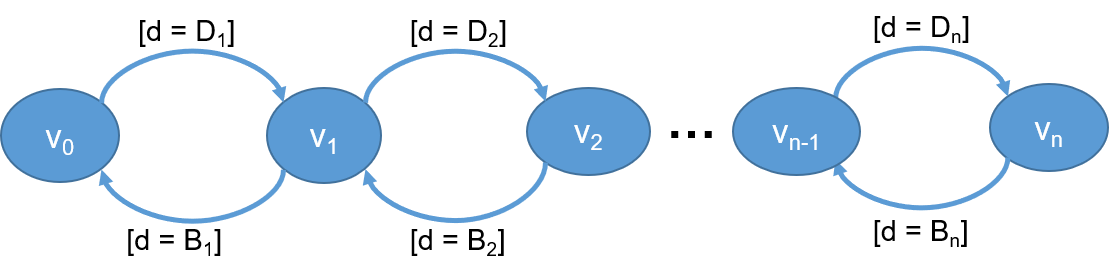}
  \caption{Automaton modelling the collision avoidance principle}
  \label{automaton}
\end{figure}

Fig.\ref{automaton} provides a scheme for the computation of $Control(d, v_t)$ in the form of an extended automaton.
 The control states correspond to speed levels $v_0, ..., v_n$.
 The transitions model instantaneous acceleration and braking steps 
 triggered by conditions involving the distance $d$ and the pre-computed bounds $B_{i}$ and $D_{i}$.
 If the control location is $v_{i}$ and the distance is equal to the minimal safe acceleration distance (i.e., $d = D_{i+1}$),
 then the automaton moves to location $v_{i+1}$ after the speed is increased to $v_{i+1}$.
 If the distance reaches the minimal safe braking distance (i.e., $d = B_{i}$),
 then the automaton moves to location $v_{i-1}$ after the speed is decreased to $v_{i-1}$.
 Given that $B_{i} = B_{i-1} + B(v_i, v_{i-1})$, 
 after braking to $v_{i-1}$ there is still enough space for safe braking.
 If none of the triggering conditions holds, 
 then the distance $d$ is such that $B_i < d < D_{i+1}$.
 The automaton stays at location $v_i$ and the speed remains unchanged.

Details about the implementation of this control principle can be found in \cite{wang2020safe}. 
In our context, the speed $v_{safe}$ is computed for nominal conditions 
considering that $v_t$ = $v_e$ - $v_a$ and that the braking and acceleration functions are defined for moderate rates. 
Hence, $v_{safe}$ = Control( $d$, $v_e$ - $v_a$).

For out-of-nominal conditions, 
 the maximal speed $v_{max}$ of the ego vehicle is $v_{max}$ = Max\{ $v ~ | ~ B_{max}(v_e) \leq d$ \},
 where $B_{max}$ is the deceleration function for some maximal deceleration rate. 
 To make sure that collision is avoided in any case, 
 the controller compares the speeds  $v_{safe}$ and $v_{max}$ and 
 when $v_{max}$ is reached, a command is issued for emergency braking.

Compared with the MPC Controller, 
 the Safe Controller can guarantee safety and efficiency 
 since at any time it chooses the speed minimizing the relative distance depending on simple criteria.
 On the contrary, the MPC Controller applies more involved computation 
 trying to estimate the future states of both vehicles according to the kinematic model of the vehicle, 
 which often requires some intelligent algorithms (e.g., genetic computation). 
 Thus, it may require computationally expensive optimization techniques. 
 On the contrary, the Safe Controller is computationally cheap and 
 can be easily implemented in real time without additional costs.

\subsection{Evaluation of the two control approaches}

\subsubsection{Experimental setting and evaluation criteria} 
\ 
\vspace{0.1cm}
\newline
We implement the MPC Controller and the Safe Controller in the Carla (version 0.9.8) simulator \cite{Dosovitskiy17}.

For the Safe Controller,  the acceleration/deceleration in the nominal setting is taken $3 ~ m/s^2$
 and the deceleration rate $12 ~ m/s^2$ in the out-of-nominal setting.
 The speed levels are from the set $\{0, 4, 8, 12,16,20,24,28,32\} (m/s)$.
 The control policy for the computation of $v_{safe}$ is based on the relative speed $v_e$ - $v_a$.

For the MPC Controller, the prediction horizon is set to 10 steps.
 The constant $d_c$ in Equation \ref{x-opt} is $20$ m.
 The weighting matrix $Q$ is set to $diag [50, 400, 1]^{T}$. 
 Time lag $\tau$ is 0.3, and control stimulus weight $r$ is set to 1.

To evaluate the efficiency of the two controllers,
 we carry out a set of comprehensive experiments performed  on a Windows 10 PC with AMD R5 3500 and NVIDIA GTX 1660 SUPER.
  We consider both nominal and out-of-nominal scenarios.
  
 \begin{enumerate}
 \item In a nominal scenario, the speed of the vehicle ahead is described by the function 
  $v_{a}(t) = A \sin(\frac{2 \pi}{T} t) + v_{a, 0}$, where $v_{a, 0}$ is taken equal to $12 ~ m/s$.
   In the experiments,  we consider three different values of $A$, i.e., $\{6, 9, 12\} ~ m/s$, 
   and three different values of  $T$, i.e., $ \{10, 20, 30\} ~ s$.
 \item In an out-of-nominal scenario, the vehicle ahead brakes suddenly and stops.
  We assume that the ego vehicle does not know when the sudden braking may occur.
 \end{enumerate}

In all scenarios the ego vehicle and the vehicle ahead move in the same lane in the same direction.
 The initial speed of the ego vehicle is set to $0 ~ m/s$. 
 The initial relative distance between the two vehicles is set to $10 ~ m$.

We check whether the MPC Controller violates safety for nominal and out-of-nominal scenarios.
 For nominal scenarios, 
 we evaluate the efficiency of the two controllers 
 with respect to the following three criteria defined for simulation time $t_{sim}$.

\begin{enumerate}
\item  Performance is measured as the ratio of  average speed of the ego vehicle 
with respect to that of the vehicle ahead, i.e.,
\begin{equation*}
\mathcal{M}_{p} = \frac{\int_0^{t_{sim}} v_{e}(t) dt}{\int_0^{t_{sim}} v_{a}(t) dt} 
\end{equation*}
 where $v_{e}$ is the speed of the ego vehicle and $v_{a}$ is the speed of the vehicle ahead.
\item Road occupancy is defined as the ratio of the space occupied by the vehicles over the total available space. 
 For our case with two vehicles, 
 we consider $1/d$ as a measure of the occupancy, where $d$ is their relative distance.
 For a simulation in time interval $[0, t_{sim}]$, it is given by the formula: 
\begin{equation*}
\mathcal{M}_{o} = \frac{1}{t_{sim}} \int_0^{t_{sim}} 1/d(t) dt
\end{equation*}
 The higher the value $\mathcal{M}_{o}$, the higher the occupancy.
 Note that the measure uses $d$,  the distance between the two vehicles without taking into account any safety margin.
\item Comfort means that the variations of acceleration are close to its average value.
 We consider that it is measured as the reciprocal of the acceleration variance, i.e., 
\begin{equation*}
\mathcal{M}_{c} = ( \frac{1}{t_{sim}}\int_0^{t_{sim}} (a(t) - \bar{a})^2 dt ) ^{-1}
\end{equation*}
where $a$ is the acceleration of the ego vehicle,  and $\bar{a} = \frac{1}{t_{sim}} \int_0^{t_{sim}} a(t) dt$.
 Note that the higher the value $\mathcal{M}_{c}$, the higher the comfort level.
\end{enumerate}

\subsubsection{Evaluation of the two controllers}
\begin{figure} [H]
  \centering
  \includegraphics[width=0.7\hsize]{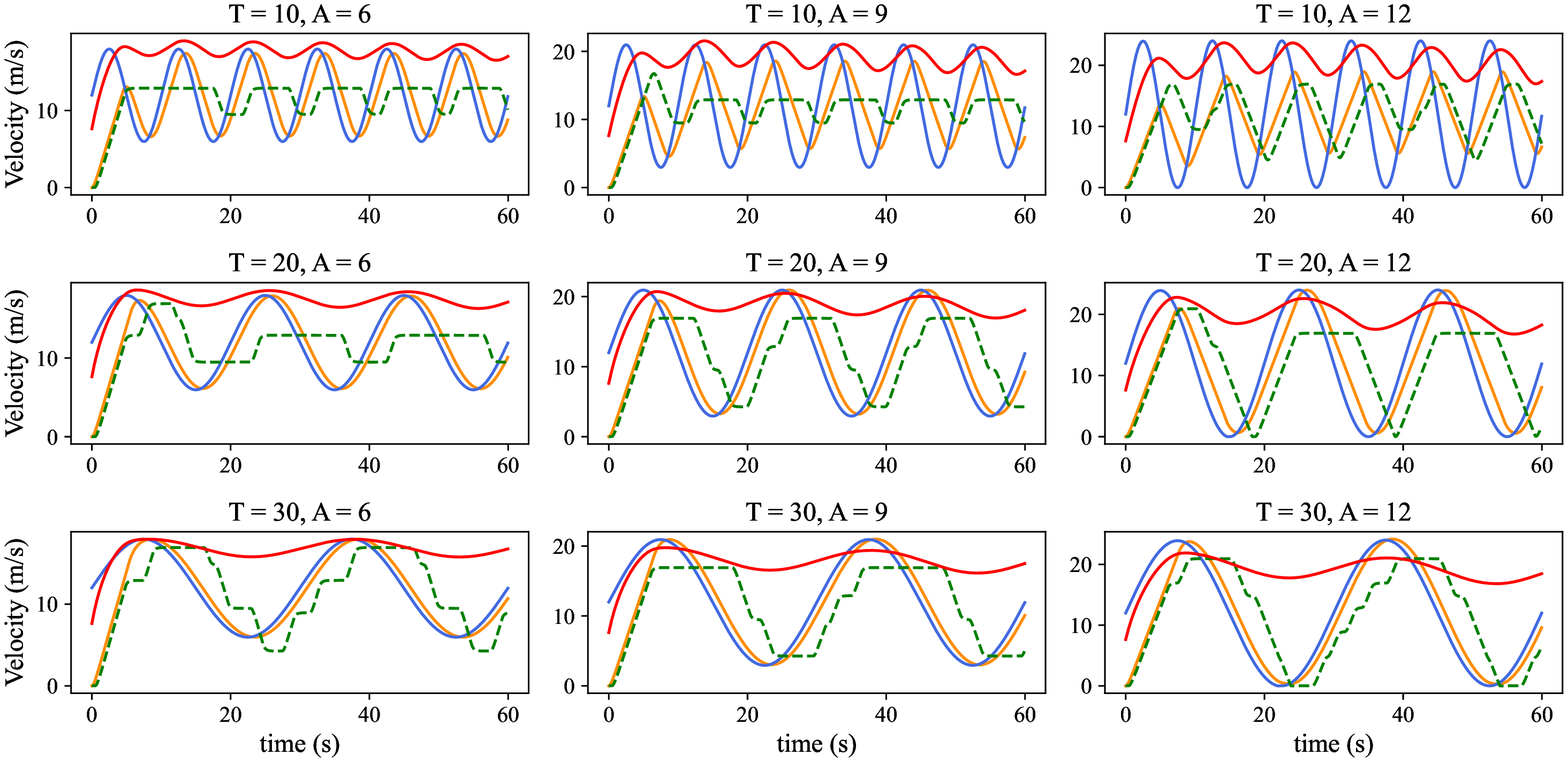}
  \caption{Speed for the Safe Controller (in green), the MPC Controller (in orange) 
  and the maximal speed $v_{max}$ (in red) for nominal scenarios. 
  Note the unsafe situations where $v_{mpc} > v_{max}$ .}
  \label{velocity-plot}
\end{figure}

\ 
\vspace{0.2cm}
\newline
\textbf{Nominal scenarios}
 Fig. \ref{velocity-plot} shows the speed of the ego vehicle for the two controllers 
 and the maximal safe speed $v_{max}$ for the nominal scenarios. 
 Note that the MPC Controller closely follows the speed of the vehicle ahead, 
 in particular when the period of the speed function increases, e.g., T = 30 s. 
 However, the speed of the MPC Controller cannot avoid unsafe situations as shown in Fig.\ref{velocity-plot}
 when the orange line ($v_{mpc}$) crosses the red line ($v_{max}$).
 On the contrary, the Safe Controller is less sensitive to speed changes of the vehicle ahead
 and allows smaller speed variation due to safety constraints.
 Despite these constraints, the performance measured as the average speed has not been sacrificed.
 The upmost part of Table.\ref{metrics-10} provides performance metrics showing that in most cases, 
 the Safe Controller produces higher average speeds than the MPC Controller and thus maintains slightly higher ratios. 
 Nonetheless, the differences increase when the amplitude of the speed function becomes larger, e.g., A = 12.

\begin{figure} [H]
  \centering
  \includegraphics[width=0.7\hsize]{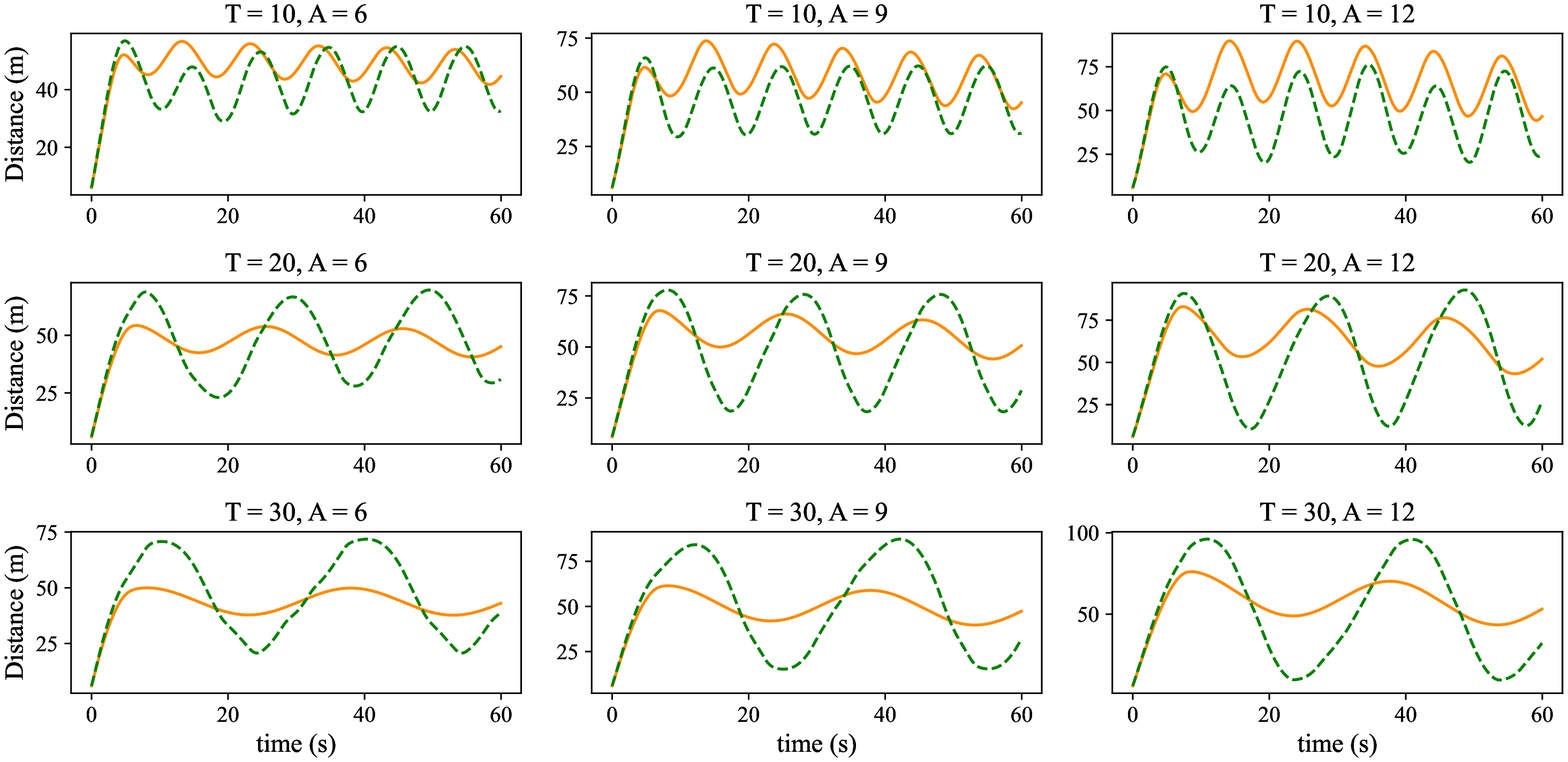}
  \caption{Relative distance for the Safe Controller (in green) and the MPC Controller (in orange) for nominal scenarios.}
  \label{distance-plot}
\end{figure}

Fig. \ref{distance-plot} provides the relative distance for the two controllers. 
 It shows that the amplitude variation for the MPC Controller is much smaller,
 especially when the period and the amplitude of the speed function are larger.
 This is because the MPC Controller favors speed tracking.
 On the contrary, the Safe Controller maintains a smaller relative distance on average than the MPC Controller,
 since its control policy focuses on distance minimization.
 This observation is confirmed by the higher occupancy metrics for the Safe Controller provided in Table.\ref{metrics-10}.

The comfort metrics provided by Table. \ref{metrics-10} show that 
 comfort for the Safe Controller is much higher than for the MPC Controller
 when the period and the amplitude of the speed function are small. 
 The reason is that the Safe Controller is less sensitive to the speed changes of the vehicle ahead 
 and avoids alternating changes of acceleration and deceleration. 
 Nonetheless, for larger periods, e.g. T = 30, the Safe Controller is less comfortable.

\vspace{-1cm}

\vspace{-1cm}
 
 \begin{table} [H]
  \centering
  \includegraphics[width=0.6\hsize]{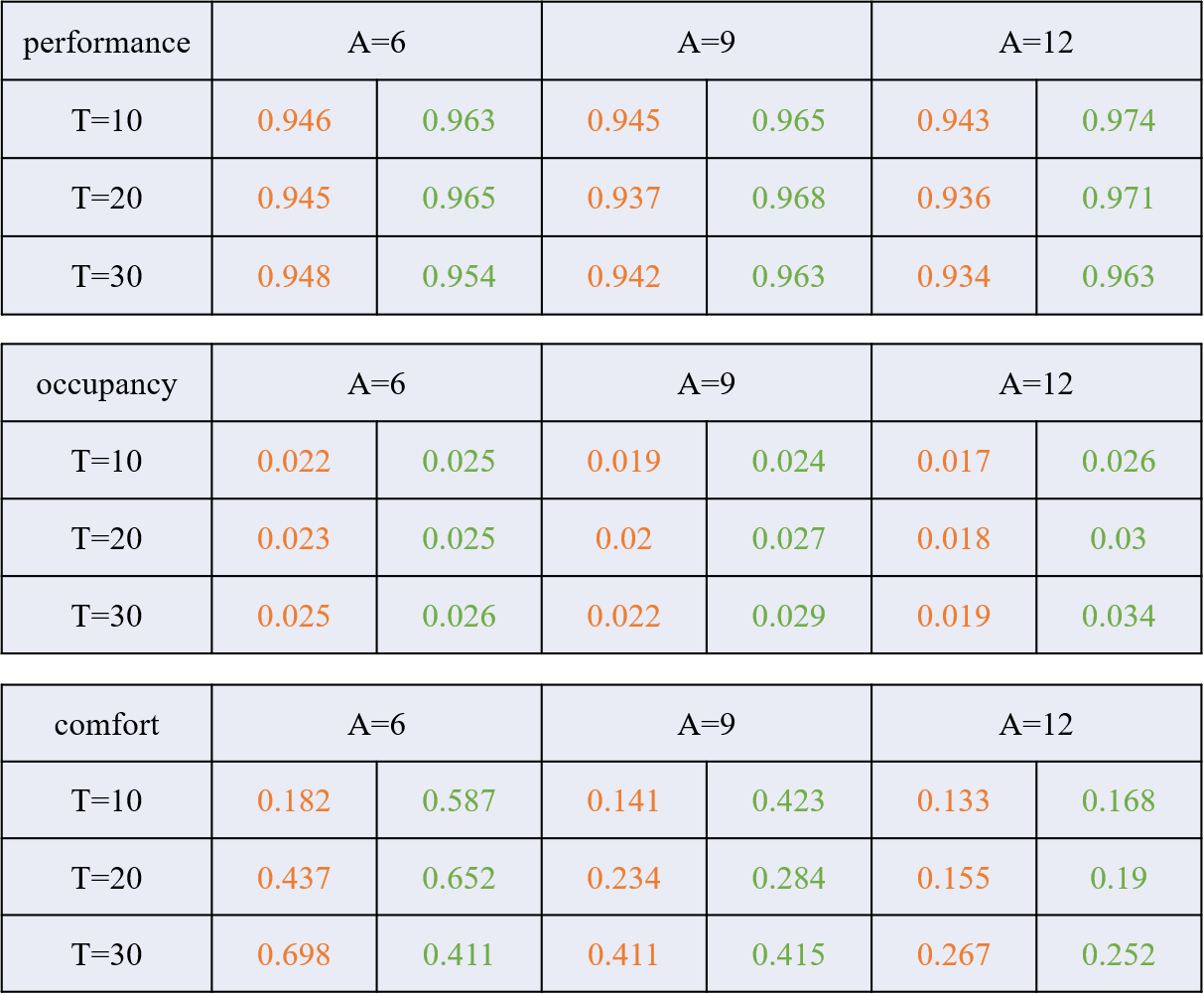}
  \caption{Efficiency metrics for the Safe Controller (in green) and the MPC Controller (in orange) for nominal scenarios 
  when the initial relative distance is 10 m.}
  \label{metrics-10}
\end{table}

\begin{figure} [H]
  \centering
  \includegraphics[width=0.7\hsize]{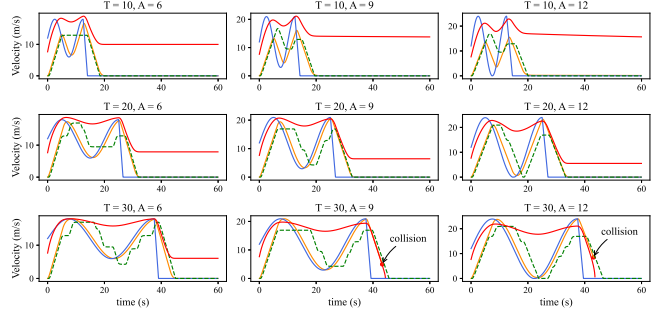}
  \caption{Speed for the Safe Controller (in green) and the MPC Controller (in orange) in out-of-nominal scenarios
   when the vehicle ahead suddenly brakes.}
  \label{velocity-break-plot}
\end{figure}

\begin{figure} [H]
  \centering
  \includegraphics[width=0.7\hsize]{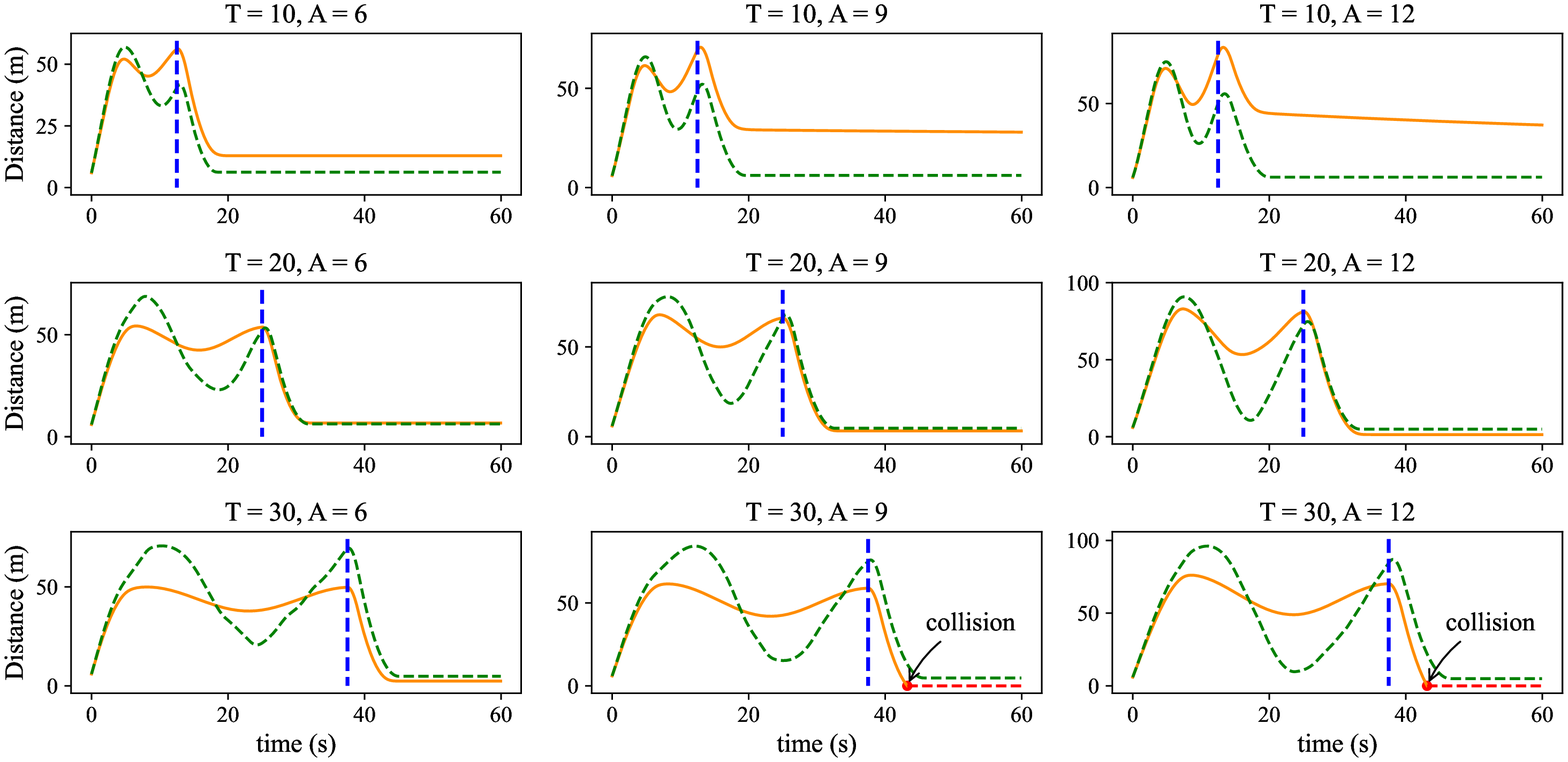}
  \caption{Relative distance for the Safe Controller (in green) and the MPC Controller (in orange)
  when the vehicle ahead suddenly brakes at the moment indicated by the dashed blue line.}
  \label{distance-break-plot}
\end{figure}

 \noindent
\textbf{Out-of-nominal scenarios}
 Fig. \ref{velocity-break-plot} and Fig. \ref{distance-break-plot} 
 show the speed and the relative distance for the emergency scenarios, respectively.
 Note that both controllers react to a sudden brake of the vehicle ahead.
 The Safe Controller always maintains a safe distance between the two vehicles,  thus avoiding collision.  
 On the contrary,  the MPC Controller is unsafe in two out of nine cases,
 in particular, when the period of the speed function becomes large.
 For instance, in Fig. \ref{distance-break-plot} 
 we can see that a collision occurs after 40 seconds when T= 30 s and A = 12 m/s 
 (the blue dashed line marks the beginning of the braking).

\vspace{-0.5cm}
\vspace{-1.5cm}

\vspace{-1cm}

\section{Hybrid collision avoidance control}
\label{sec:hybrid}

\subsection{Design and implementation of the Hybrid Controller}

The above comparative evaluation confirms that 
 in most cases the MPC Controller is slightly better in terms of performance and occupancy, 
 while it fails to be safe for out-of-nominal scenarios when the vehicle ahead abruptly stops.
 As expected the Safe Controller preserves safety in all scenarios and assures more comfortable driving.

We have explained the design principle of the Hybrid Controller in the Introduction (Fig.\ref{combination-arch}). 
 It consists of the Safe Controller and the MPC Controller running in parallel 
 and a Switch deciding which one of the control stimuli takes effect. 
 The Switch produces the target speed of the ego vehicle taking care that it never exceeds the maximal safe speed $v_{max}$
 computed as a function of the distance $d$ and the maximal deceleration of the ego vehicle.
 For constant braking rate $a_{max}$, we have $v_{max} = (2\cdot a_{max} \cdot d) ^{1/2}$.

In the Hybrid Controller under the constraint $v \leq v_{max}$,
 the Switch selects the highest speed between $v_{mpc}$ and $v_{safe}$ so as to achieve the best performance.
 Hence, it continuously applies the following rules to select the target speed $v$ of the Hybrid Controller: \\
 \textbf{if} $v_{safe}  \leq v_{mpc} \leq v_{max} $ \textbf{then} $v := v_{mpc}$ \\
 \textbf{else if}  $v_{mpc} \leq v_{safe}$ \textbf{then} $v := v_{safe}$ \\
 \textbf{else if} $v_{max} \leq v_{mpc}$ \textbf{then}  $v := v_{max}$

Notice that in all cases the rules prevent the target speed $v$ from exceeding $v_{max}$. 
 In nominal conditions it can happen that $v_{mpc} \leq v_{safe}$. 
 This is the case when the distance $d$ is large enough and the speed of the vehicle ahead is decreasing. 
 The speed $v_{safe}$ can exceed $v_{mpc}$ as the Safe Controller focuses on minimizing the relative distance 
 while the MPC Controller tracks the speed of the vehicle ahead.

An important difference from similar works dealing with hybrid controllers \cite{provebecorrectacc,2017Adaptive} 
 is that our Safe Controller contributes not only to safety but also to a large extent to performance 
 and comfort and even in some cases to improve occupancy. 
 This observation is confirmed by experimental results provided in the next section.
 Notice that this hybrid control principle can be applied 
 by replacing in our architecture the MPC Controller by other adaptive controllers, 
 including controllers based on machine learning \cite{learning2016}.

\subsection{Evaluation of the Hybrid Controller}

We consider both nominal and out-of-nominal scenarios as before and adopt the same experimental settings. 
 For nominal scenarios, we compare the efficicency of the three controllers. 
 Furthermore, for out-of-nominal scenarios we consider additional braking rates of the vehicle ahead.

\begin{figure} [!htb]
  \centering
  \includegraphics[width=0.7\hsize]{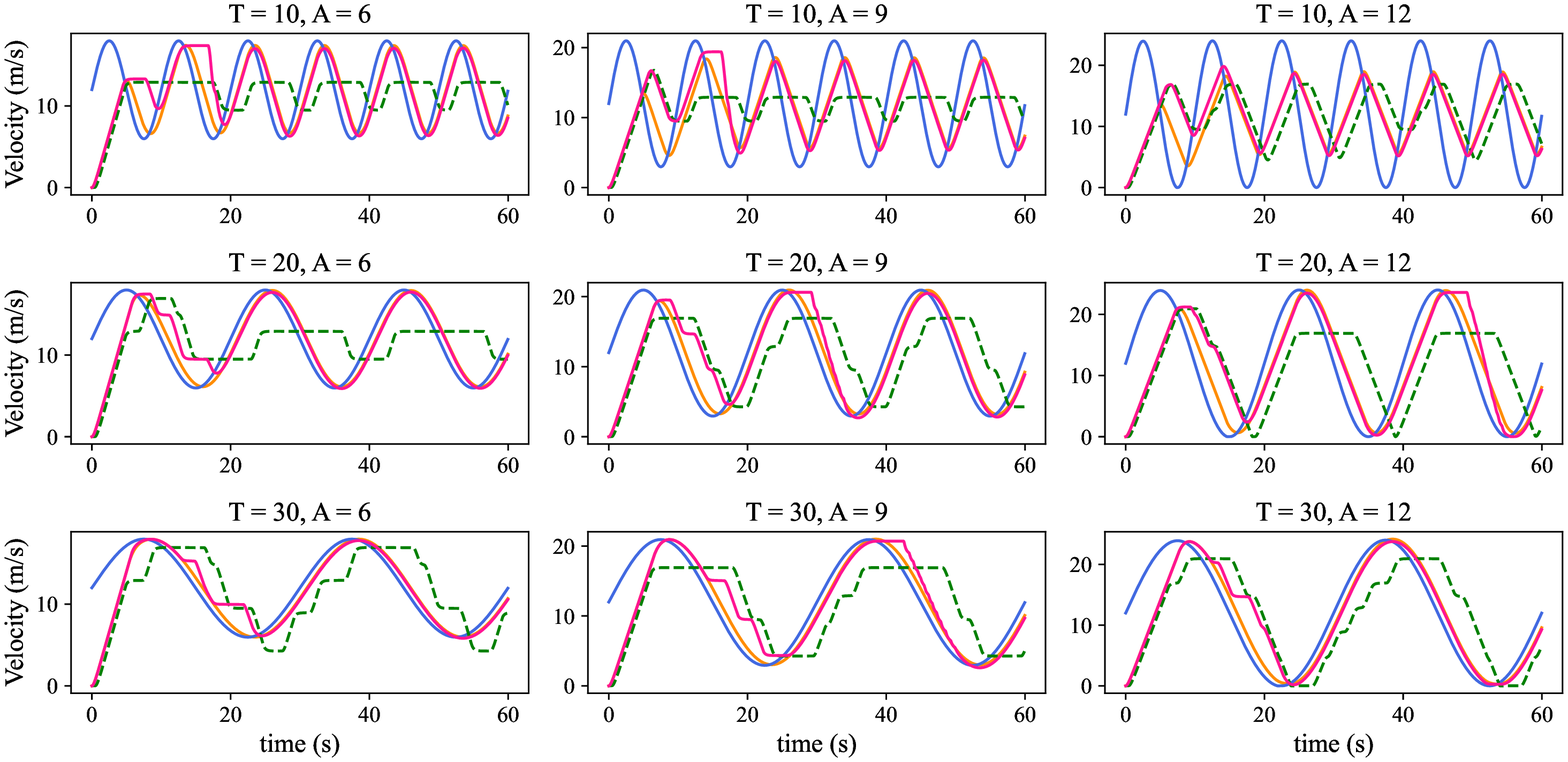}
  \caption{Speed for the Hybrid Controller (in purple), the MPC Controller (in orange), the Safe Controller (in green) 
  when the speed of the vehicle ahead is a sinusoidal function (in blue).}
  \label{combine-velocity-plot}
\end{figure}

\vspace{0.25cm}
\noindent
\textbf{Nominal scenarios}
 Fig.\ref{combine-velocity-plot} depicts the speed of the ego vehicle for the three controllers in nominal scenarios. 
 It shows that the Hybrid Controller can also track the speed of the vehicle ahead closely, taking the best from the two controllers.
 Notice that the purple line can be above the orange line, 
 for instance during simulation time around 20 seconds when T = 30 and A =9.
 The upmost part of Table.\ref{metrics-hybrid-10} provides results comparing the performance of the three controllers.  
 Note that the Hybrid Controller outperforms the two other controllers.

\begin{figure} [!htb]
  \centering
  \includegraphics[width=0.7\hsize]{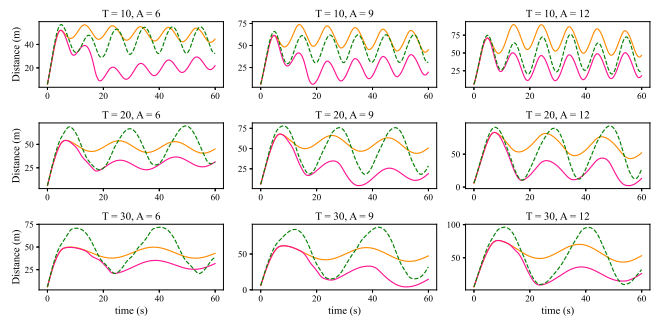}
  \caption{Relative distance for the Hybrid Controller (in purple), 
  the MPC Controller (in orange) and the Safe Controller (in green) 
  when the speed of the vehicle ahead is a sinusoidal function.}
  \label{combine-distance-plot}
\end{figure}

Fig.\ref{combine-distance-plot} depicts the relative distance for the three controllers. 
 As expected the distance maintained by the Hybrid Controller is in general smaller 
 taking advantage of the strength of the Safe Controller for minimizing the relative distance. 
 This is also shown in the middle part of Table \ref{metrics-hybrid-10},
 which compares the occupancy for the three controllers. 
 The Hybrid Controller achieves higher occupancy than the other two controllers.

The comfort metrics provided in the bottom part of Table \ref{metrics-hybrid-10},
 show that when the period and the amplitude of the speed function are small,  
 the Safe Controller produces the most comfortable driving policies. 
 While when they become larger,
 the Hybrid Controller is better than the Safe Controller and is slightly outperformed by the MPC Controller.

Table \ref{time-percentage} shows time percentages corresponding to the application 
 by the Hybrid Controller of the MPC policy, the nominal safe policy and the out-of-nominal safe policy.
 We can see that the MPC Controller contributes more than the other two,
 while the contribution of the Safe Controller is non-negligible.
 The maximal safe speed $v_{max}$ is applied to a very small percentage of cases to ensure safety.

\begin{figure} [!htb]
  \centering
  \includegraphics[width=0.7\hsize]{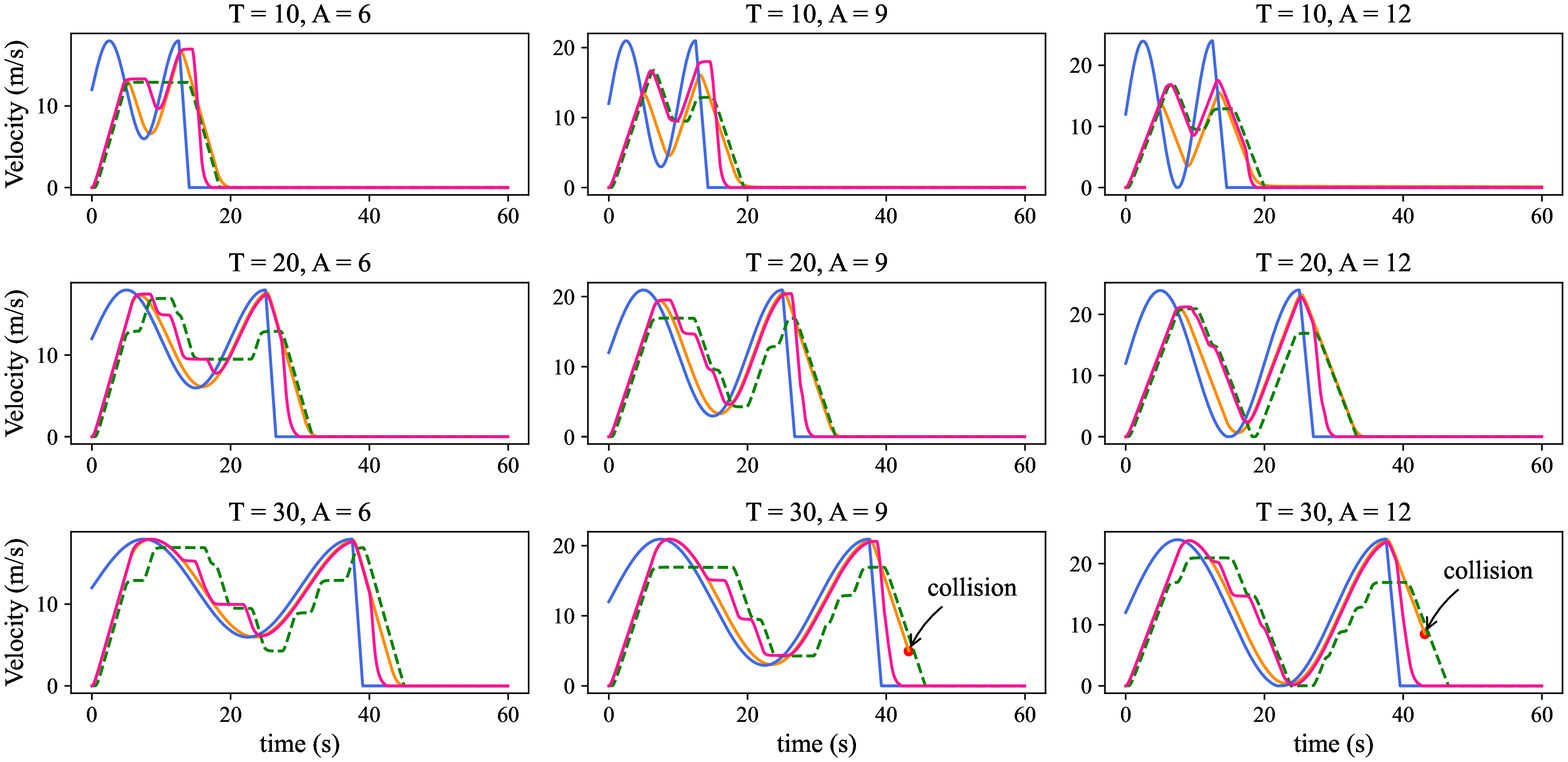}
  \caption{Speed of the ego vehicle for the Hybrid Controller (in purple), 
  the MPC Controller (in orange) and the Safe Controller (in green) 
  when the vehicle ahead brakes with rate $12 ~ m/s^{2}$.}
  \label{combine-velocity-break-plot}
\end{figure}

\begin{figure} [!htb]
  \centering
  \includegraphics[width=0.7\hsize]{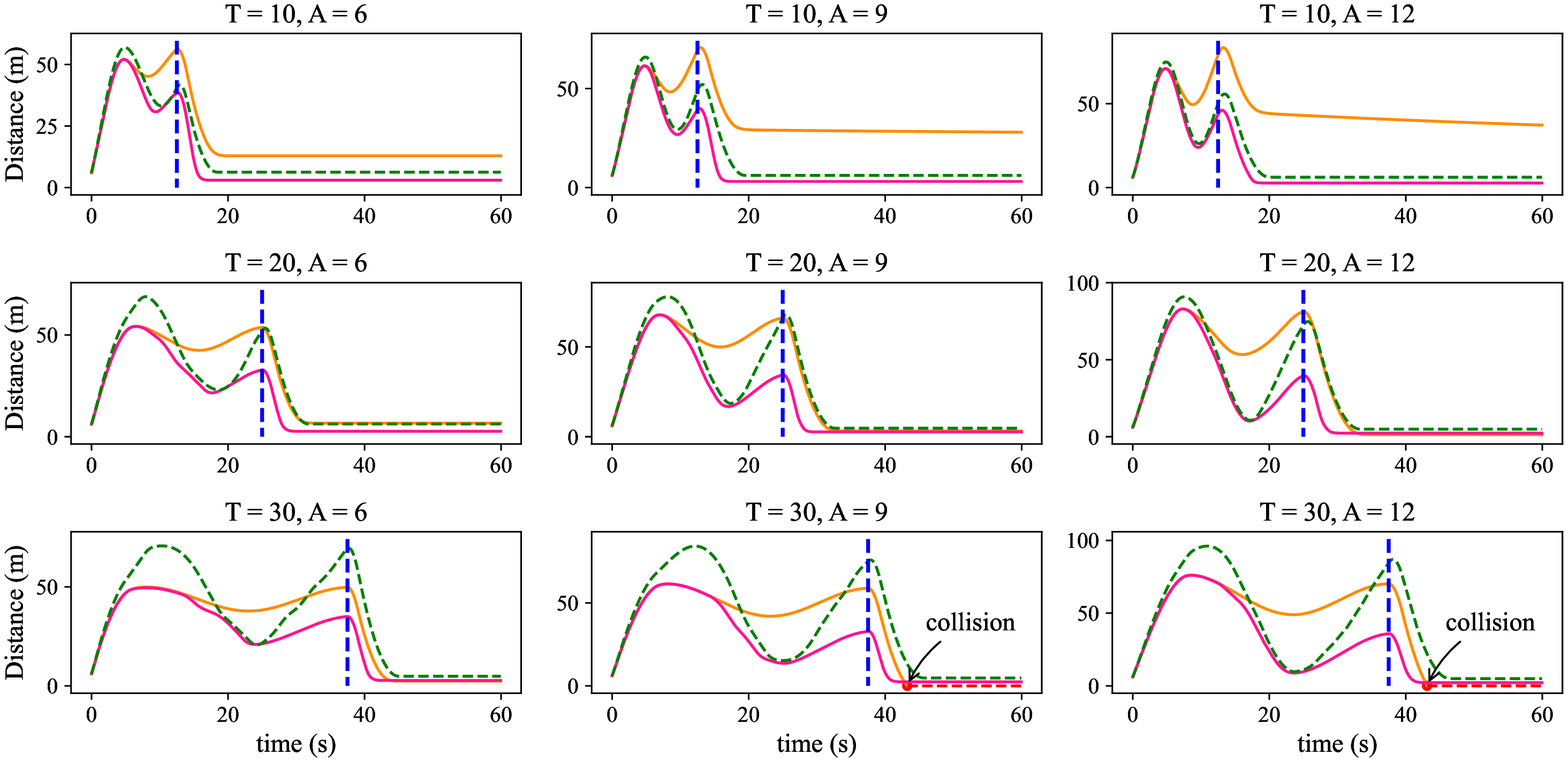}
  \caption{Relative distance for the Hybrid Controller (in purple),  
  the MPC Controller (in orange) and the Safe Controller (in green)  
  when the vehicle ahead brakes with rate $12 ~ m/s^{2}$.}
  \label{combine-distance-break-plot}
\end{figure}

\vspace{0.25cm}
\noindent
\textbf{Out-of-nominal scenarios}
Fig.\ref{combine-velocity-break-plot} and Fig.\ref{combine-distance-break-plot}
 provide results for the three controllers in the out-of-nominal scenarios 
 where the vehicle ahead suddenly brakes with rate $12 ~ m/s^2$. 
 Note that the MPC Controller becomes unsafe for increasing amplitude and period of the speed of the vehicle ahead. 
 There are two out of nine settings where the MPC control policy results in collision.
 In Annex A,  additional experiments for braking rates $4 ~ m/s^2$ and $8 ~ m/s^2$ are provided.
 %
 They show that the MPC Controller is safe for all scenarios with braking rate $4 ~ m/s^2$ 
 while it is fails to be safe in one out of nine scenarios with braking rate $8 ~ m/s^2$.
 
\vspace{-1.5cm}

\vspace{-1.5cm}

 \begin{table} [H]
  \centering
  \includegraphics[width=0.7\hsize]{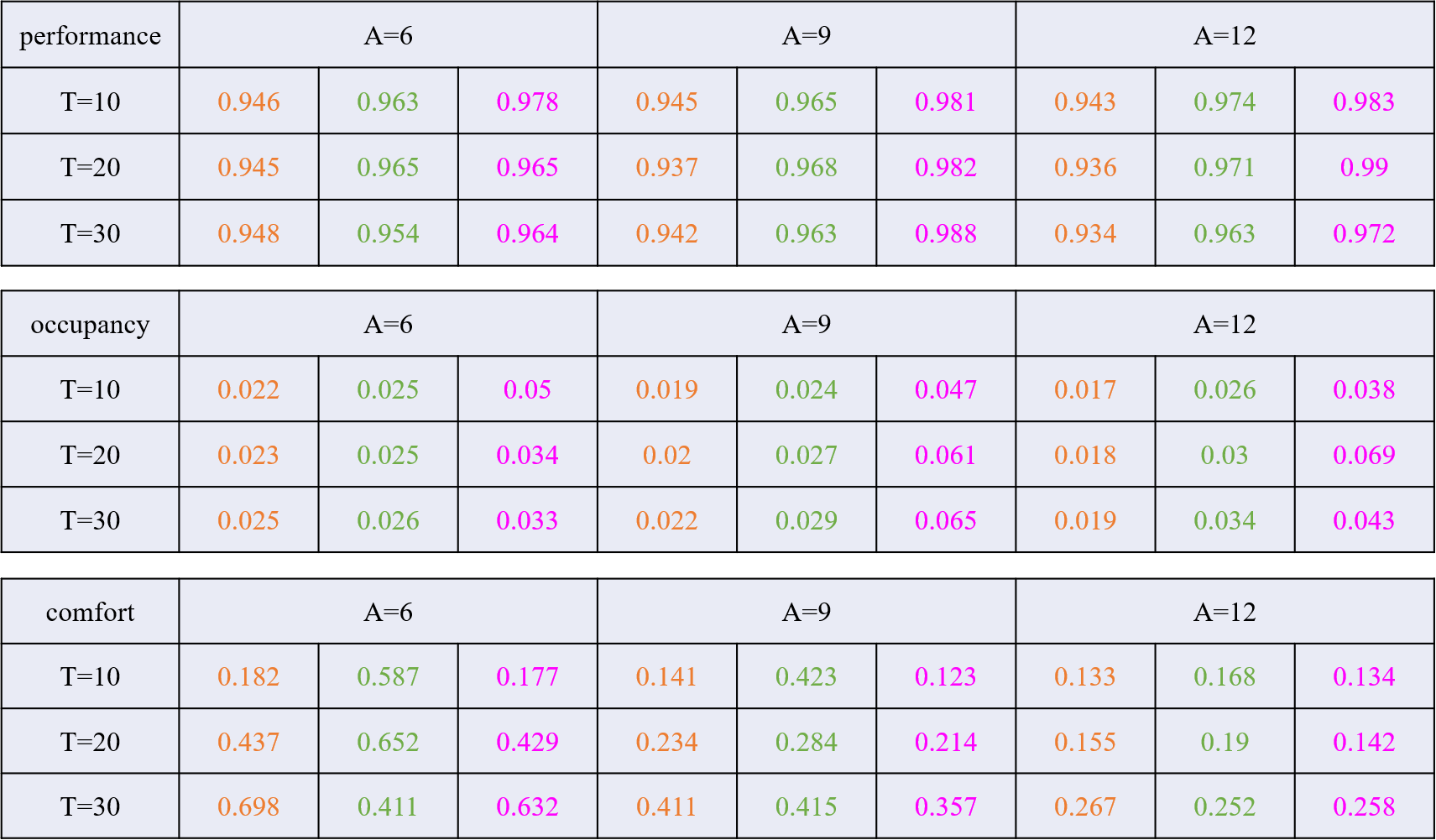}
  \caption{Efficiency metrics for the Hybrid Controller (in purple), 
  the Safe Controller (in green) and the MPC Controller (in orange) for nominal scenarios. }
  \label{metrics-hybrid-10}
\end{table}
\vspace{-1.5cm}

\begin{table} [H]
  \centering
  \includegraphics[width=0.7\hsize]{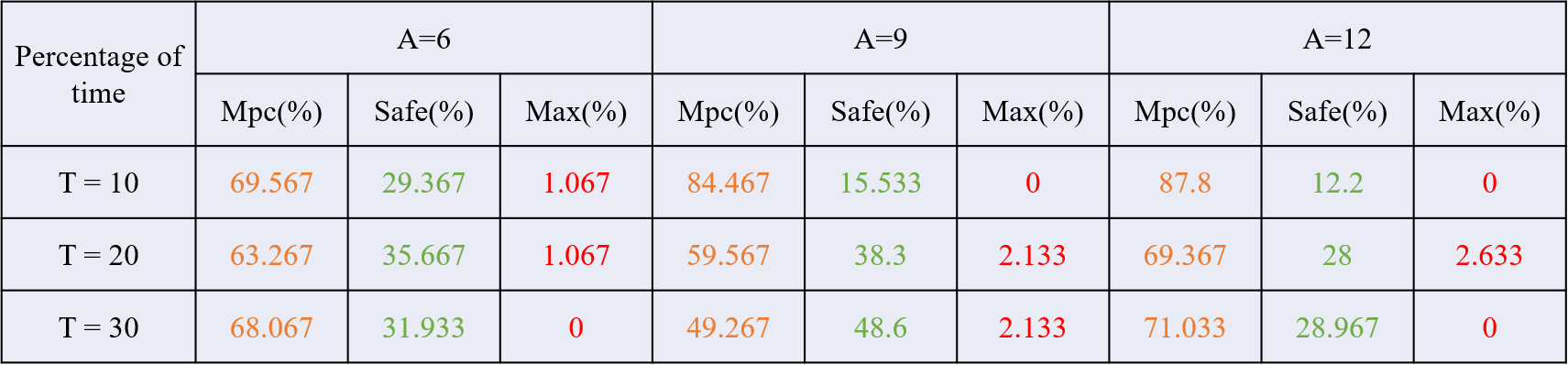}
  \caption{Percentage of time in use for the Safe Controller, 
  the MPC Controller and the maximal safe speed $v_{max}$ for nominal scenarios. }
  \label{time-percentage}
\end{table}

\section{Conclusion and discussion}
\label{sec:conclusion}

We propose a method for building a hybrid safe-by-construction and efficient collision avoidance controller. 
 The controller integrates a  MPC Controller,  a  discrete Safe Controller 
 and a Switch that combines the outputs of the two controllers to generate stimuli that are safe and efficient. 
 We show experimentally that the Hybrid Controller besides guaranteeing safety, ensures high efficiency 
 because it "takes the best" from each one of the integrated controllers. 
 The MPC Controller seeks policies that reduce both the relative speed and the relative distance 
 while in nominal scenarios the Safe Controller seeks minimization of the relative distance.
 We show that this hybrid control policy ensures a very good efficiency 
 measured by three criteria: performance, road occupancy and comfort. 

We adopt a pragmatic and progressive methodology 
 based on the comparative evaluation of the two constituent controllers for both  nominal and out-of-nominal scenarios.  
 The evaluation provides a good insight on the merits of the respective control principles which motivates the design of the Hybrid Controller. 
 The experimental results confirm the feasibility and the practical relevance of hybrid controllers  for safe and efficient driving. 

A key and original lesson from our results is that  the Safe Controller is not simply a monitor that takes over in critical situations. 
 It also significantly contributes to efficiency applying a control policy that nicely complements the MPC policy. 
 The experimental results show that the interplay between the dynamics of discrete and continuous controllers 
 pursuing complementary objectives can be surprisingly rich. 
 Its study may lead to more elaborated and enhanced hybrid policies. 
 In future work we will  investigate our hybrid control  principle 
 by replacing the MPC Controller with other types of  adaptive controllers, e.g., machine-learning-based controllers \cite{learning2016}.

\bibliographystyle{splncs03}
\bibliography{main}

\clearpage
\appendix


\section*{Annex A}

\begin{figure} [!htb]
  \centering
  \includegraphics[width=0.7\hsize]{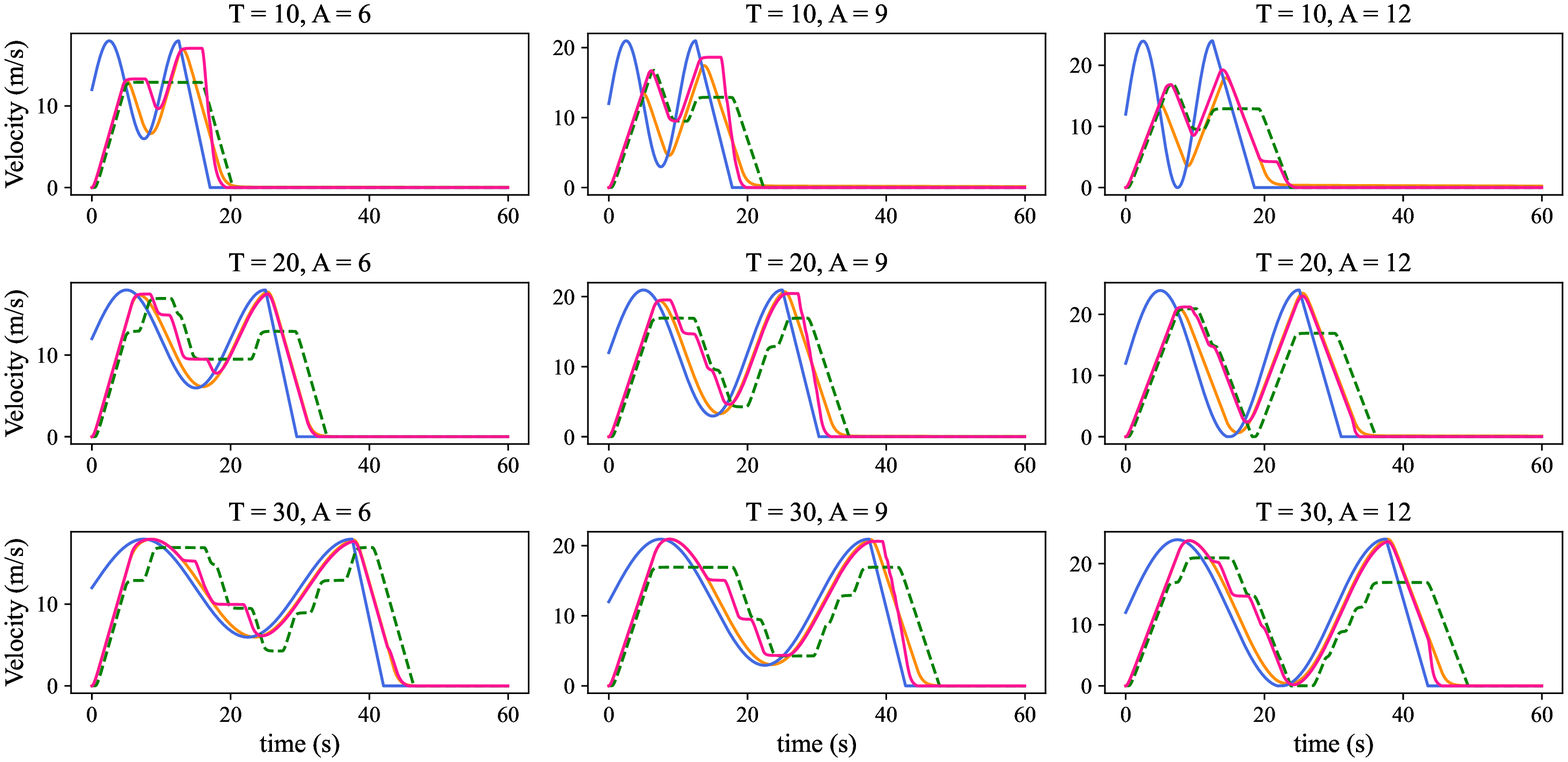}
  \caption{Speed of the ego vehicle for the Hybrid Controller (in purple), 
  the MPC Controller (in orange) and the Safe Controller (in green) 
  when the vehicle ahead brakes with rate $4 ~ m/s^{2}$.}
  \label{combine-velocity-plot-m4}
\end{figure}
\vspace{-0.5cm}

\begin{figure} [!htb]
  \centering
  \includegraphics[width=0.7\hsize]{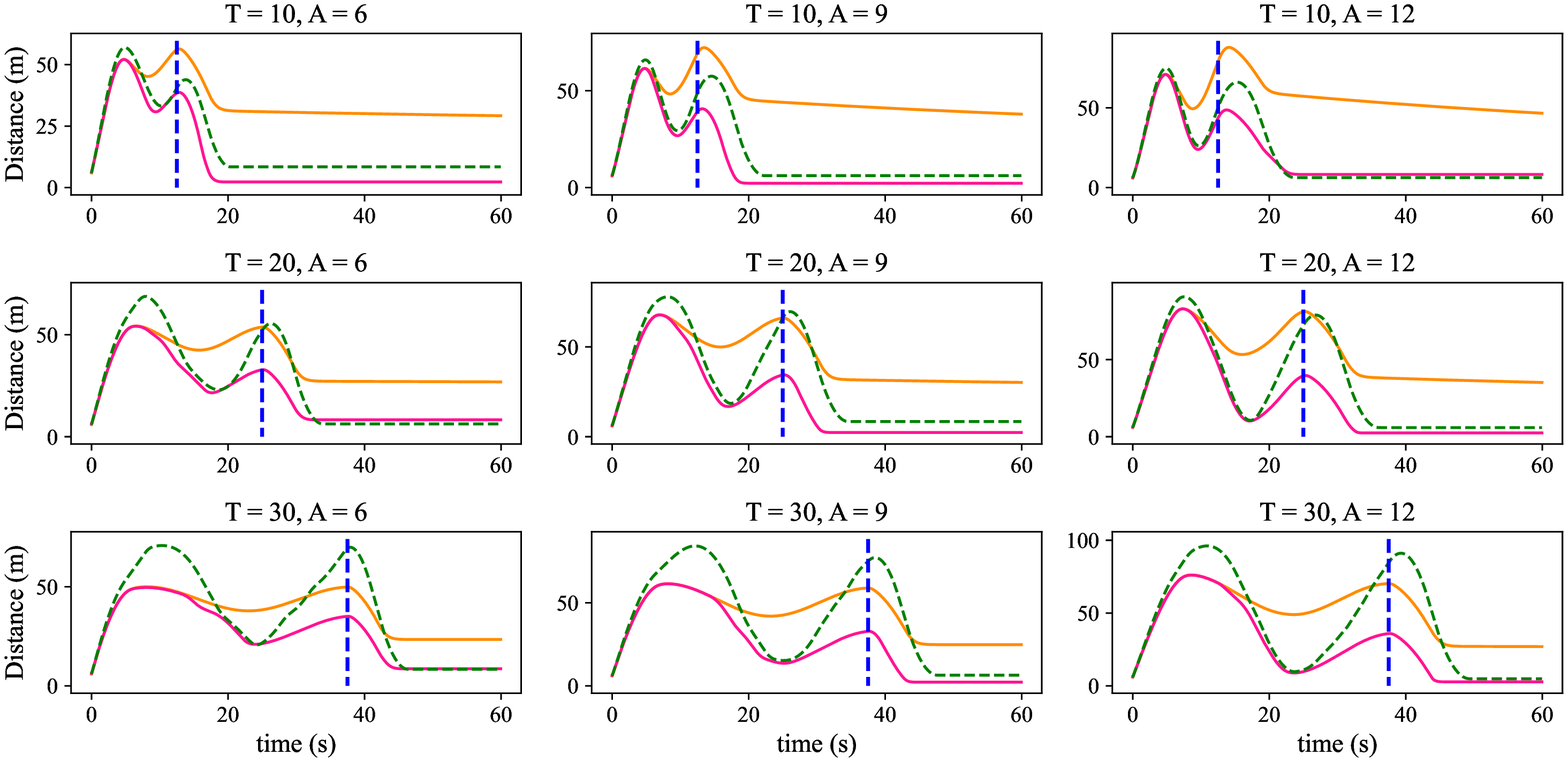}
  \caption{Relative distance for the Hybrid Controller (in purple), 
  the MPC Controller (in orange) and the Safe Controller (in green)
  when the vehicle ahead brakes with rate $4 ~ m/s^{2}$.}
  \label{combine-distance-plot-m4}
\end{figure}

\begin{figure} [!htb]
  \centering
  \includegraphics[width=0.7\hsize]{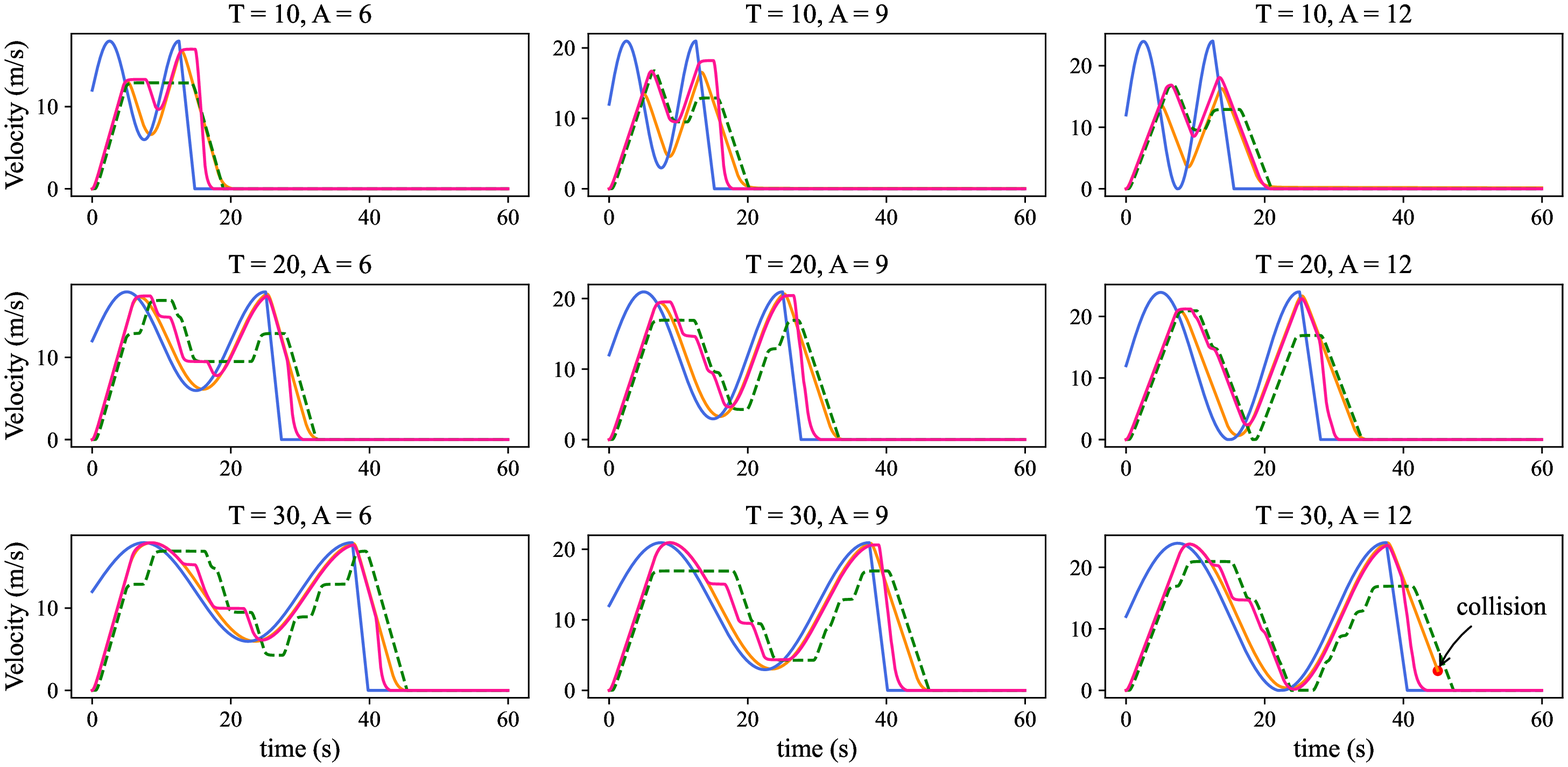}
  \caption{Speed of the ego vehicle for the Hybrid Controller (in purple),  
  the MPC Controller (in orange) and the Safe Controller (in green) 
  when the vehicle ahead brakes with rate $8 ~ m/s^{2}$.}
  \label{combine-velocity-break-plot-m8}
\end{figure}

\begin{figure} [!htb]
  \centering
  \includegraphics[width=0.7\hsize]{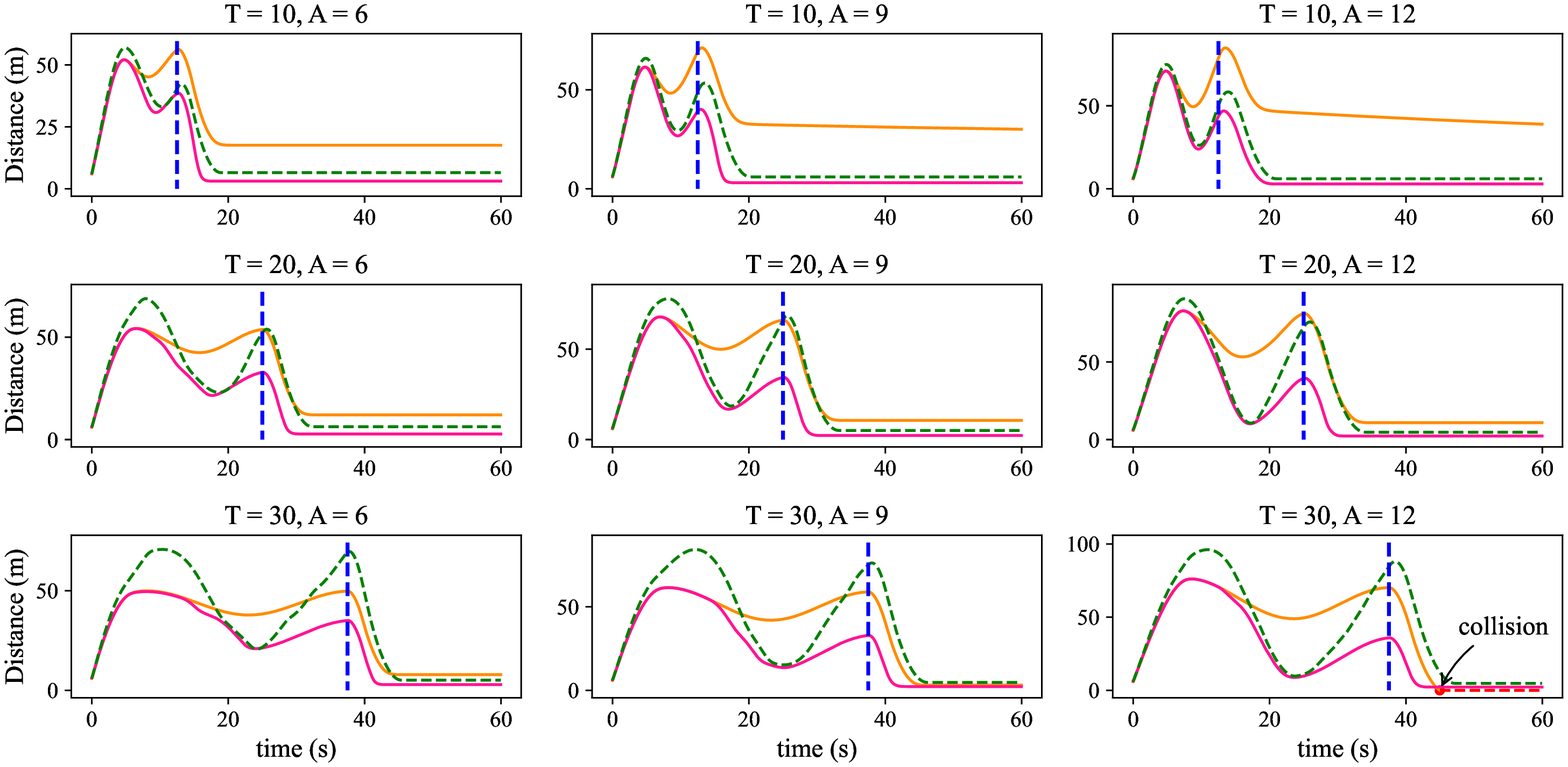}
  \caption{Relative distance for the Hybrid Controller (in purple), 
  the MPC Controller (in orange) and the Safe Controller (in green) 
  when the vehicle ahead brakes with rate $8 ~ m/s^{2}$.}
  \label{combine-distance-break-plot-m8}
\end{figure}

\end{document}